# Measuring the Similarity of Sentential Arguments in Dialog

**Amita Misra, Brian Ecker, and Marilyn A. Walker**
University of California Santa Cruz
Natural Language and Dialog Systems Lab
1156 N. High. SOE-3
Santa Cruz, California, 95064, USA
`amitamisra|becker|maw@soe.ucsc.edu`

## Abstract

When people converse about social or political topics, similar arguments are often paraphrased by different speakers, across many different conversations. Debate websites produce curated summaries of arguments on such topics; these summaries typically consist of lists of sentences that represent frequently paraphrased propositions, or labels capturing the essence of one particular aspect of an argument, e.g. `Morality` or `Second Amendment`. We call these frequently paraphrased propositions ARGUMENT FACETS. Like these curated sites, our goal is to induce and identify argument facets across multiple conversations, and produce summaries. However, we aim to do this automatically. We frame the problem as consisting of two steps: we first extract sentences that express an argument from raw social media dialogs, and then rank the extracted arguments in terms of their similarity to one another. Sets of similar arguments are used to represent argument facets. We show here that we can predict ARGUMENT FACET SIMILARITY with a correlation averaging 0.63 compared to a human topline averaging 0.68 over three debate topics, easily beating several reasonable baselines.

## 1 Introduction

When people converse about social or political topics, similar arguments are often paraphrased by different speakers, across many different conversations. For example, consider the dialog excerpts in Fig. 1 from the 89K sentences about gun control in the IAC 2.0 corpus of online dialogs (Abbott et al., 2016). Each of the sentences **S1** to **S6** provide different linguistic realizations of the same proposition namely that *Criminals will have guns even if gun ownership is illegal.*

| |
|---|
| **S1**: To inact a law that makes a crime of illegal gun ownership has no effect on criminal ownership of guns.. |
| **S2**: Gun free zones are zones where criminals will have guns because criminals will not obey the laws about gun free zones. |
| **S3**: Gun control laws do not stop criminals from getting guns. |
| **S4**: Gun control laws will not work because criminals do not obey gun control laws! |
| **S5**: Gun control laws only control the guns in the hands of people who follow laws. |
| **S6**: Gun laws and bans are put in place that only affect good law abiding free citizens. |

Figure 1: Paraphrases of the *Criminals will have guns* facet from multiple conversations.

Debate websites, such as Idebate and ProCon produce curated summaries of arguments on the gun control topic, as well as many other topics.[1][2] These summaries typically consist of lists, e.g. Fig. 2 lists eight different aspects of the gun control argument from Idebate. Such manually curated summaries identify different linguistic realizations of the same argument to induce a set of common, repeated, aspects of arguments, what we call ARGUMENT FACETS. For example, a curator might identify sentences **S1** to **S6** in Fig. 1 with a label to represent the facet that *Criminals will have guns even if gun ownership is illegal.*

Like these curated sites, we also aim to induce and identify facets of an argument across multiple conversations, and produce summaries of all the different facets. However our aim is to do this automatically, and over time. In order to simplify the problem, we focus on SENTENTIAL ARGUMENTS, single sentences that clearly express a particular argument facet in dialog. We aim to use SENTENTIAL ARGUMENTS to produce extractive summaries of online dialogs about current social and political topics. This paper extends our previous work which frames our goal as consisting of two tasks (Misra et al., 2015; Swanson et al., 2015).

---

[1] See `http://debatepedia.idebate.org/en/index.php/Debate:_Gun_control`,
[2] See `http://gun-control.procon.org/`

| Pro Arguments |
|---|
| **A1**: The only function of a gun is to kill. |
| **A2**: The legal ownership of guns by ordinary citizens inevitably leads to many accidental deaths. |
| **A3**: Sports shooting desensitizes people to the lethal nature of firearms. |
| **A4**: Gun ownership increases the risk of suicide. |
| **Con Arguments** |
| **A5**: Gun ownership is an integral facet of the right to self defense. |
| **A6**: Gun ownership increases national security within democratic states. |
| **A7**: Sports shooting is a safe activity. |
| **A8**: Effective gun control is not achievable in democratic states with a tradition of civilian gun owership. |

Figure 2: The eight facets for Gun Control on IDebate, a curated debate site.

- **Task1: Argument Extraction**: How can we extract sentences from dialog that clearly express a particular argument facet?
- **Task2: Argument Facet Similarity**: How can we recognize that two sentential arguments are semantically similar, i.e. that they are different linguistic realizations of the same facet of the argument?

**Task1** is needed because social media dialogs consist of many sentences that either do not express an argument, or cannot be understood out of context. Thus sentences that are useful for inducing argument facets must first be automatically identified. Our previous work on Argument Extraction achieved good results, (Swanson et al., 2015), and is extended here (Sec. 2).

**Task2** takes pairs of sentences from Task1 as input and then learns a regressor that can predict Argument Facet Similarity (henceforth **AFS**). Related work on argument mining (discussed in more detail in Sec. 4) defines a finite set of facets for each topic, similar to those from Idebate in Fig. 2.[3] Previous work then labels posts or sentences using these facets, and trains a classifier to return a facet label (Conrad et al., 2012; Hasan and Ng, 2014; Boltuzic and Šnajder, 2014; Naderi and Hirst, 2015), *inter alia*. However, this simplification may not work in the long term, both because the sentential realizations of argument facets are propositional, and hence graded, and because facets evolve over time, and hence cannot be represented by a finite list.

In our previous work on AFS, we developed an AFS regressor for predicting the similarity of **human-generated labels** for summaries of dialogic arguments (Misra et al., 2015). We collected 5 human summaries of each dialog, and then used the Pyramid tool and scheme to annotate sentences from these summaries as contributors to (paraphrases of) a particular facet (Nenkova and Passonneau, 2004). The Pyramid tool requires the annotator to provide a human readable label for a collection of contributors that realize the same propositional content. The AFS regressor operated on pairs of human-generated labels from Pyramid summaries of different dialogs about the same topic. In this case, facet identification is done by the human summarizers, and collections of similar labels represent an argument facet. We believe this is a much easier task than the one we attempt here of training an AFS regressor on automatically extracted raw sentences from social media dialogs. The contributions of this paper are:

- We develop a new corpus of sentential arguments with gold-standard labels for AFS.
- We analyze and improve our argument extractor, by testing it on a much larger dataset. We develop a larger gold standard corpus for ARGUMENT QUALITY (AQ).
- We develop a regressor that can predict AFS on extracted sentential arguments with a correlation averaging **0.63** compared to a human topline of **0.68** for three debate topics.[4]

## 2 Corpora and Problem Definition

Many existing websites summarize the frequent, and repeated, facets of arguments about current topics, that are linguistically realized in different ways, across many different social media and debate forums. For example, Fig. 2 illustrates the eight facets for gun control on IDebate. Fig. 3 illustrates a different type of summary, for the death penalty topic, from ProCon, where the argument facets are called out as the "Top Ten Pros and Cons" and given labels such as `Morality`, `Constitutionality` and `Race`. See the top of Fig. 3. The bottom of Fig. 3 shows how each facet is then elaborated by a paragraph for both its Pro and Con side: due to space we only show the summary for the `Morality` facet here.

These summaries are curated, thus one would

---

[3] See also the facets in Fig. 3 below from `ProCon.org`.

[4] Both the AQ and the AFS pair corpora are available at `nlds.soe.ucsc.edu`.

![ProCon.org Death Penalty facets screenshot]

Figure 3: Facets of the death penalty debate as curated on ProCon.org

not expect that different sites would call out the exact same facets, or even that the same type of labels would be used for a specific facet. As we can see, ProCon (Fig. 3) uses one word or phrasal labels, while IDebate (Fig. 2) describes each facet with a sentence. Moreover, these curated summaries are not produced for a particular topic once-and-for-all: the curators often reorganize their summaries, drawing out different facets, or combining previously distinct facets under a single new heading. We hypothesize that this happens because new facets arise over time. For example, it is plausible that for the gay marriage topic, the facet that *Gay marriage is a civil rights issue* came to the fore only in the last ten years.

Our long-term aim is to produce summaries similar to these curated summaries, but automatically, and over time, so that as new argument facets arise for a particular topic, we can identify them. We begin with three debate topics, gun control (38102 posts), gay marriage (22425 posts) and death penalty (5283 posts), from the Internet Argument Corpus 2.0 (Abbott et al., 2016). We first need to create a large sample of high quality sentential arguments (**Task1** above) and then create a large sample of paired sentential arguments in order to train the model for AFS (**Task2** above).

### 2.1 Argument Quality Data

We extracted all the sentences for all of the posts in each topic to first create a large corpus of topic-sorted sentences. See Table 1.

We started with the Argument Quality (**AQ**) re-

| Topic | Original | Rescored | Sampled | AQ #N (%) |
|-------|----------|----------|---------|-----------|
| GC    | 89,722   | 63,025   | 2140    | 1887 (88%) |
| DP    | 17,904   | 11,435   | 1986    | 1520 (77%) |
| GM    | 51,543   | 40,306   | 2062    | 1745 (85%) |

Table 1: Sentence count in each domain. Sampled bin range > 0.55 and number of sentential arguments (high AQ) after annotation. GC=Gun Control, DP=Death Penalty, GM=Gay Marriage.

gressor from Swanson et al. (2015), which gives a score to each sentence. The AQ score is intended to reflect how easily the speaker's argument can be understood from the sentence without any context. Easily understandable sentences are assumed to be prime candidates for producing extractive summaries. In Swanson et al. (2015), the annotators rated AQ using a continuous slider ranging from hard (0.0) to easy to interpret (1.0). We refined the Mechanical Turk task to elicit new training data for AQ as summarized in Table 1. Fig. 8 in the appendix shows the HIT we used to collect new AQ labels for sentences, as described below.

We expected to to apply Swanson's AQ regressor to our sample completely "out of the box". However, we first discovered that many sentences given high AQ scores were very similar, while we need a sample that realizes many **diverse** facets. We then discovered that some extracted sentential arguments were not actually high quality. We hypothesized that the diversity issue arose primarily because Swanson's dataset was filtered using high PMI n-grams. We also hypothesized that the quality issue had not surfaced because Swanson's sample was primarily selected from sentences marked with the discourse connectives *but*, *first*, *if*, and *so*. Our sample (Original column of Table 1) is much larger and was not similarly filtered.

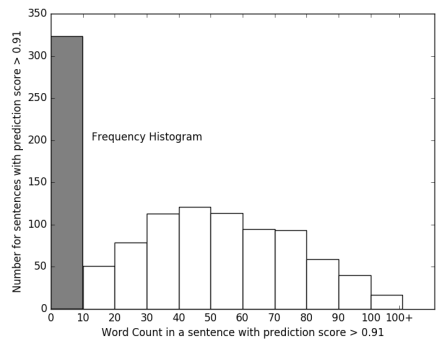

Figure 4: Word count distribution for argument quality prediction scores > 0.91 for Swanson's original model.

Fig. 4 plots the distribution of word counts for sentences from our sample that were given an AQ score > 0.91 by Swanson's trained AQ regressor. The first bin shows that many sentences with

278

less than 10 words are predicted to be high quality, but many of these sentences in our data consisted of only a few elongated words (e.g. HA-HAHAHA...). The upper part of the distribution shows a large number of sentences with more than 70 words with a predicted AQ > 0.91. We discovered that most of these long sentences are multiple sentences without punctuation. We thus refined the AQ model by removing duplicate sentences, and rescoring sentences without a verb and with less than 4 dictionary words to AQ = 0. We then restricted our sampling to sentences between 10 and 40 tokens, to eliminate run-on sentences and sentences without much propositional content. We did not retrain the regressor, rather we resampled and rescored the corpus. See the Rescored column of Table 1. After removing the two tails in Fig. 4, the distribution of word counts is almost uniform across bins of sentences from length 10 to 40.

As noted above, the sample in Swanson et al. (2015) was filtered using PMI, and PMI contributes to AQ. Thus, to end up with a diverse set of sentences representing many facets of each topic, we decided to sample sentences with lower AQ scores than Swanson had used. We binned the sentences based on predicted AQ score and extracted random samples across bins ranging from .55–1.0, in increments of .10. Then we extracted a smaller sample and collected new AQ annotations for gay marriage and death penalty on Mechanical Turk, using the definitions in Fig. 8 (in the appendix). See the Sampled column of Table 1. We pre-selected three annotators using a qualifier that included detailed instructions and sample annotations. A score of 3 was mapped to a yes and scores of 1 or 2 mapped to a no. We simplified the task slightly in the HIT for gun control, where five annotators were instructed to select a yes label if the sentence clearly expressed an argument (score 3), or a no label otherwise (score 1 or 2).

We then calculated the probability that the sentences in each bin were high quality arguments using the resulting AQ gold standard labels, and found that a threshhold of predicted AQ > 0.55 maintained both diversity and quality. See Fig. 9 in the appendix. Table 1 summarizes the results of each stage of the process of producing the new AQ corpus of 6188 sentences (Sampled and then annotated). The last column of Table 1 shows that gold standard labels agree with the rescored AQ regressor between 77% and 88% of the time.

## 2.2 Argument Facet Similarity Data

The goal of **Task2** is to define a similarity metric and train a regression model that takes as input two sentential arguments and returns a scalar value that predicts their similarity(AFS). The model must reflect the fact that similarity is graded, e.g. the same argument facet may be repeated with different levels of explicit detail. For example, sentence A1 in Fig. 2 is similar to the more complete argument, *Given the fact that guns are weapons—things designed to kill—they should not be in the hands of the public*, which expresses both the premise and conclusion. Sentence A1 leaves it up to the reader to infer the (obvious) conclusion.

| |
|---|
| **S7**: Since there are gun deaths in countries that have banned guns, the gun bans did not work. |
| **S8**: It is legal to own weapons in this country, they are just tightly controlled, and as a result we have far less gun crime (particularly where it's not related to organised crime). |
| **S9**: My point was that the theory that more gun control leaves people defenseless does not explain the lower murder rates in other developed nations. |

Figure 5: Paraphrases of the *Gun ownership does not lead to higher crime* facet of the Gun Control topic across different conversations.

Our approach to **Task2** draws strongly on recent work on semantic textual similarity (STS) (Agirre et al., 2013; Dolan and Brockett, 2005; Mihalcea et al., 2006). STS measures the degree of semantic similarity between a pair of sentences with values that range from 0 to 5. Inspired by the scale used for STS, we first define what a facet is, and then define the values of the AFS scale as shown in Fig. 10 in the appendix (repeated from Misra et al. (2015) for convenience). We distinguish AFS from STS because: (1) our data are so different: STS data consists of descriptive sentences whereas our sentences are argumentative excerpts from dialogs; and (2) our definition of facet allows for sentences that express opposite stance to be realizations of the same facet (AFS = 3) in Fig. 10.

Related work has primarily used entailment or semantic equivalence to define argument similarity (Habernal and Gurevych, 2015; Boltuzic and Šnajder, 2015; Boltuzic and Šnajder, 2015; Habernal et al., 2014). We believe the definition of AFS given in Fig. 10 will be more useful in the long run than semantic equivalence or entailment, because two arguments can only be contradictory if they are about the same facet. For example, consider that sentential argument **S7** in Fig. 5 is anti gun-control, while sentences **S8** and **S9** are pro gun-control. Our annotation guidelines label them with the same facet, in a similar way to how the curated summaries on ProCon provides both a Pro and Con side for each facet. See Fig. 3.

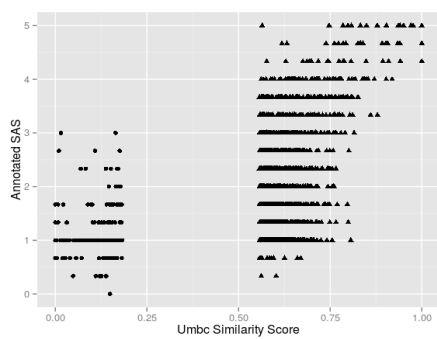

Figure 6: The distribution of AFS scores as a function of UMBC STS scores for gun control sentences.

In order to efficiently collect annotations for AFS, we want to produce training data pairs that are more likely than chance to be the same facet (scores 3 and above as defined in Fig. 10). Similar arguments are rare with an all-pairs matching protocol, e.g. in ComArg approximately 67% of the annotations are "not a match" (Boltuzic and Šnajder, 2014). Also, we found that Turkers are confused when asked to annotate similarity and then given a set of sentence pairs that are almost all highly dissimilar. Annotations also cost money. We therefore used UMBC STS (Han et al., 2013) to score all potential pairs.[5] To foreshadow, the plot in Fig. 6 shows that this pre-scoring works: (1) the lower quadrant of the plot shows that STS $< .20$ corresponds to the lower range of scores for AFS; and (2) the lower half of the left hand side shows that we still get many arguments that are low AFS (values below 3) in our training data.

We selected 2000 pairs in each topic, based on their UMBC similarity scores, which resulted in lowest UMBC scores of 0.58 for GM, 0.56 for GC and 0.58 for DP. To ensure a pool of diverse arguments, a particular sentence can appear in at most ten pairs. MT workers took a qualification test with definitions and instructions as shown in Fig. 10. Sentential arguments with sample AFS annotations were part of the qualifier. The 6000 pairs were made available to our three most reliable pre-qualified workers. The last row of Table 3 reports the human topline for the task, i.e. the average pairwise $r$ across all three workers. Interestingly, the Gay marriage topic ($r = 0.60$) is more difficult for human annotators than either Death Penalty ($r = 0.74$) or Gun Control ($r = 0.69$).

---

[5]This is an off-the-shelf STS tool from University of Maryland Baltimore County available at swoogle.umbc.edu/SimService/.

## 3 Argument Facet Similarity

Given the data collected above, we defined a supervised machine learning experiment with AFS as our dependent variable. We developed a number of baselines using off the shelf tools. Features are grouped into sets and discussed in detail below.

### 3.1 Feature Sets

**NGRAM cosine.** Our primary baseline is an ngram overlap feature. For each argument, we extract the unigrams, bigrams and trigrams, and then calculate the cosine similarity between two texts represented as vectors of their ngram counts.

**Rouge.** Rouge is a family of metrics for comparing the similarity of two summaries (Lin, 2004), which measures overlapping units such as continuous and skip ngrams, common subsequences, and word pairs. We use all the rouge f-scores from the pyrouge package. Our analysis shows that rouge_s*_f_score correlates most highly with AFS.[6]

**UMBC STS.** We consider STS, a measure of the semantic similarity of two texts (Agirre et al., 2012), as another baseline, using the UMBC STS tool. Fig. 6 illustrates that in general, STS is rough approximation of AFS. It is possible that our selection of data for pairs for annotation using UMBC STS either improves or reduces its performance.

**Google Word2Vec.** Word embeddings from word2vec (Mikolov et al., 2013) are popular for expressing semantic relationships between words, but using word embeddings to express entire sentences often requires some compromises. In particular, averaging word2vec embeddings for each word may lose too much information in long sentences. Previous work on argument mining has developed methods using word2vec that are effective for clustering similar arguments (Habernal and Gurevych, 2015; Boltuzic and Šnajder, 2015) Other research creates embeddings at the sentence level using more advanced techniques such as Paragraph Vectors (Le and Mikolov, 2014).

We take a more direct approach in which we use the word embeddings directly as features. For each sentential argument in the pair, we create a 300-dimensional vector by filtering for stopwords and punctuation and then averaging the word embeddings from Google's word2vec model for the remaining words.[7] Each dimension of the 600 dimensional concatenated averaged vector is used directly as a feature. In our experiments, this concatenation method greatly outperforms cosine similarity (Table 2, Table 3). Sec. 3.3 discusses properties of word embeddings that may yield these performance differences.

---

[6]https://pypi.python.org/pypi/pyrouge/
[7]https://code.google.com/archive/p/word2vec/

**Custom Word2Vec.** We also create our own 300-dimensional embeddings for our dialogic domain using the Gensim library (Řehůřek and Sojka, 2010), with default settings, and a very large corpus of user-generated dialogic content. This includes the corpus described in Sec. 2 (929, 206 forum posts), an internal corpus of 1, 688, 639 tweets on various topics, and a corpus of 53, 851, 542 posts from Reddit.[8]

**LIWC category and Dependency Overlap.** Both dependency structures and the Linguistics Inquiry Word Count (LIWC) tool have been useful in previous work (Pennebaker et al., 2001; Somasundaran and Wiebe, 2009; Hasan and Ng, 2013). We develop a novel feature set that combines LIWC category and dependency overlap, aiming to capture a generalized notion of concept overlap between two arguments, i.e. to capture the hypothesis that classes of content words such as affective processes or emotion types are indicative of a shared facet across pairs of arguments.

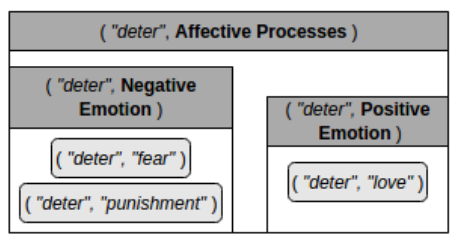

Figure 7: LIWC Generalized Dep. tuples

We create partially generalized LIWC dependency features and count overlap normalized by sentence length across pairs, building on previous work (Joshi and Penstein-Rosé, 2009). Stanford dependency features (Manning et al., 2014) are generalized by leaving one dependency element lexicalized, replacing the other word in the dependency relation with its LIWC category and by removing the actual dependency type (nsubj, dobj, etc.) from the triple. This creates a tuple of (*"governor token", LIWC category of dependent token*). We call these simplified LIWC dependencies.

Fig. 7 illustrates the generalization process for three LIWC simplified dependencies, *("deter", "fear"), ("deter", "punishment"),* and *("deter", "love")*. Because LIWC is a hierarchical lexicon, two dependencies may share many generalizations or only a few. Here, the tuples with dependent tokens *fear* and *punishment* are more closely related because their shared generalization include both *Negative Emotion* and *Affective Processes*, but the tuples with dependent tokens *fear* and *love* have a less similar relationship, because they only share the *Affective Processes* generalization.

---
[8]One month sample https://www.reddit.com/r/datasets/comments/3bxlg7/i_have_every_publicly_available_reddit_comment

### 3.2 Machine Learning Regression Results

We randomly selected 90% of our annotated pairs to use for nested 10-fold cross-validation, setting aside 10% for qualitative analysis of predicted vs. gold-standard scores. We use Ridge Regression (RR) with l2-norm regularization and Support Vector Regression (SVR) with an RBF kernel from scikit-learn (Pedregosa et al., 2011). Performance evaluation uses two standard measures, Correlation Coefficient ($r$) and Root Mean Squared Error (RMSE). A separate inner cross-validation within each fold of the outer cross-validation is used to perform a grid search to determine the hyperparameters for that outer fold. The outer cross-validation reports the scoring metrics.

**Simple Ablation Models.** We first evaluate simple models based on a single feature using both RR and SVR. Table 2, Rows 1, 2, and 3 show the baseline results: UMBC Semantic Textual Similarity (STS), Ngram Cosine, and Rouge. Surprisingly, the UMBC STS measure does not perform as well as Ngram Cosine for Death Penalty and Gay Marriage. LIWC dependencies (Row 4) perform similarly to Rouge (Row 3) across topics. Cosine similarity for the custom word2vec model (Row 5) performs about as well or better than ngrams across topics, but cosine similarity using the Google model (Row 6) performs worse than ngrams for all topics except Death Penalty. Interestingly our custom Word2Vec models perform significantly better than the Google word2vec models for Gun Control and Gay Marriage, with both much higher $r$ and lower RMSE, while performing identically for Death Penalty.

**Feature Combination Models.** Table 3 shows the results of testing feature combinations to learn which ones are complementary. Since SVR consistently performs better than RR, we use SVR only. Significance is calculated using paired t-tests between the RMSE values across folds. We paired Ngrams separately with LIWC and ROUGE to evaluate if the combination is significant. Ngram+Rouge (Row 1) is significantly better than Ngram for Gun Control and Death Penalty ($p < .01$), and Gay Marriage ($p = .03$).

|    |               | **Gun Control** |      |      |      | **Gay Marriage** |      |      |      | **Death Penalty** |      |      |      |
|    |               | **RR** |   | **SVR** |   | **RR** |   | **SVR** |   | **RR** |   | **SVR** |   |
| ID | Features | $r$ | RMSE | $r$ | RMSE | $r$ | RMSE | $r$ | RMSE | $r$ | RMSE | $r$ | RMSE |
|----|-----------|-----|------|-----|------|-----|------|-----|------|-----|------|-----|------|
| 1  | UMBC      | 0.49 | 0.90 | 0.50 | 0.90 | 0.16 | 0.90 | 0.21 | 0.90 | 0.21 | 1.16 | 0.20 | 1.20 |
| 2  | Ngram     | 0.46 | 0.91 | 0.46 | 0.92 | 0.24 | 0.88 | 0.24 | 0.91 | 0.23 | 1.16 | 0.24 | 1.18 |
| 3  | Rouge     | 0.52 | 0.88 | 0.57 | 0.86 | 0.22 | 0.89 | 0.26 | 0.90 | 0.39 | 1.10 | **0.40** | **1.11** |
| 4  | LIWC dependencies | 0.50 | 0.89 | **0.59** | **0.85** | **0.27** | **0.88** | 0.26 | 0.90 | 0.34 | 1.12 | 0.40 | 1.12 |
| 5  | CustomW2Vec Cosine | 0.47 | 0.91 | 0.52 | 0.89 | 0.22 | 0.89 | 0.25 | 0.90 | 0.29 | 1.14 | 0.30 | 1.16 |
| 6  | GoogleW2Vec Cosine | 0.40 | 0.94 | 0.47 | 0.93 | 0.16 | 0.90 | 0.20 | 0.92 | 0.29 | 1.14 | 0.30 | 1.16 |

Table 2: Results for predicting AFS with individual features using Ridge Regression (RR) and Support Vector Regression (SVR) with 10-fold Cross-Validation on the 1800 training items for each topic.

| ID | Feature Combinations with SVR | **Gun Control** | | **Gay Marriage** | | **Death Penalty** | |
|    |                               | $r$ | RMSE | $r$ | RMSE | $r$ | RMSE |
|----|-------------------------------|-----|------|-----|------|-----|------|
| 1 | Ngram- Rouge | 0.59 | 0.85 | 0.29 | 0.89 | 0.40 | 1.11 |
| 2 | Ngram- LIWC dependencies | 0.61 | 0.83 | 0.34 | 0.88 | 0.43 | 1.10 |
| 3 | Ngram- LIWC dependencies- Rouge | 0.64 | 0.80 | 0.38 | 0.86 | 0.49 | 1.05 |
| 4 | Ngram- LIWC dependencies- Rouge-UMBC | 0.65 | 0.79 | 0.40 | 0.86 | 0.50 | 1.05 |
| 5 | CustomW2Vec Concatenated vectors | 0.71 | 0.72 | 0.48 | 0.80 | 0.56 | 0.99 |
| 6 | GoogleW2Vec Concatenated vectors | 0.71 | 0.72 | 0.50 | 0.79 | 0.57 | 0.98 |
| 7 | Ngram- LIWC dependencies- Rouge- UMBC- CustomW2Vec Concatenated vectors | 0.73 | 0.70 | 0.54 | 0.77 | 0.62 | 0.93 |
| 8 | Ngram- LIWC dependencies- Rouge- UMBC- GoogleW2Vec Concatenated vectors | **0.73** | **0.70** | **0.54** | **0.77** | **0.63** | **0.92** |
| 9 | **HUMAN TOPLINE** | 0.69 | | 0.60 | | 0.74 | |

Table 3: Results for feature combinations for predicting AFS, using Support Vector Regression (SVR) with 10-fold Cross-Validation on the 1800 training items for each topic.

Ngram+LIWC (Row 2) is significantly better than Ngram for Gun Control, and Death Penalty ($p < .01$). Thus both Rouge and LIWC provide complementary information to Ngrams.

Our best result using our hand-engineered features is a combination of LIWC, Rouge, and Ngrams (Row 3). Interestingly, adding UMBC STS (Row 4) gives a small but significant improvement ($p < 0.01$ for gun control; $p = 0.07$ for gay marriage). Thus we take Ngrams, LIWC, Rouge, and UMBC STS (Row 4) as our best hand-engineered model across all topics with a correlation of 0.65 for gun control, 0.50 for death penalty and 0.40 for gay marriage. This combination is significantly better than the baselines for Ngram baseline ($p < .01$), UMBC STS ($p <= .02$) and Rouge ($p < .01$) for all three topics.

We then further combine the hand-engineered features (Row 4) with the Google Word2Vec features (Row 6), creating the model in Row 8. A paired t-test between RMSE values from each cross-validation fold for each model (Row 4 vs. Row 8 and Row 6 vs. Row 8) shows that the our hand-engineered features are complementary to Word2Vec, and their combination yields a model significantly better than either model alone ($p <$ .01). We note that although the custom word2vec model performs much better for gun control and gay marriage when using cosine, it actually performs slightly but significantly ($p = .05$) worse when using concatenation with hand-engineered features. This may simply be due to the size of the training data, i.e. the Google model used nearly twice as much training data, while our domain-specific word2vec model achieves comparable performance to the Google model with much less training data.

### 3.3 Analysis and Discussion

Although it is common to translate word embeddings into single features or reduced feature sets for similarity through the use of clustering (Habernal and Gurevych, 2015) or cosine similarity (Boltuzic and Šnajder, 2015), we show that it is possible to improve results by directly combining word embeddings with hand-engineered features. In our task, sentences were limited to a maximum of 40 tokens in order to encourage single-facet sentences, but this may have provided an additional benefit by allowing us to average word embeddings while still preserving useful signal.

Our results also demonstrate that using concate-

282

| ID | Argument 1 | Argument 2 | STS | Ngram | Rouge | LIWC dep | W2Vec | AFS | MT AFS |
|---|---|---|---|---|---|---|---|---|---|
| GC1 | You say that gun control must not be effective because the study's conclusions about gun control were inconclusive. | You're right that gun control isn't about guns, however, but 'control' is a secondary matter, a means to an end. | 1.82 | 2.56 | 2.22 | 1.53 | **1.40** | 1.5 | **1** |
| DP2 | I don't feel as strongly about the death penalty as I feel about the abortion rights debate since I can relate to the desire for vengeance that people feel. | Well I, as creator of this debate, think that there should not be a death penalty. | 1.82 | 2.38 | 2.07 | **1.29** | 1.44 | 1.24 | **1.33** |
| GC3 | They do not have the expressed, enumerated power to pass any law regarding guns in the constitution. | Which passed the law requireing "smart guns", if they ever become available (right now they do not exist). | 1.74 | 1.83 | 2.67 | 1.50 | 1.82 | **1.88** | **2.0** |
| GM4 | Technically though marriage is not discrimination, because gays are still allowed to marry the opposite sex. | Everyone has the right to marry someone of the opposite sex, and with gay marriage, everyone will have the right to marry someone of the same AND opposite sex. | 1.76 | 2.09 | 1.68 | 2.00 | **2.23** | 2.06 | **2.33** |
| GM5 | If the state wants to offer legal protections and benefits to straight married couples, it cannot constitutionally refuse equal protections to gay ones. | Same-sex couples are denied over 1,000 benefits, rights, and protections that federal law affords to married, heterosexual couples, as well as hundreds of such protections at the state level. | 1.77 | 1.91 | 1.77 | 2.66 | **3.56** | 3.72 | **3.33** |
| DP6 | In addition, it is evident that the death penalty does not deter murder rates. | BUT it is not apparent that death penalty lower crime rate. | 2.03 | 2.31 | 3.71 | 2.21 | **3.84** | 3.95 | **4.0** |
| DP7 | Living in jail for life costs less money then the death penalty. | Morality aside, no evidence of deterrence aside, the death penalty costs more than life imprisonment. | 1.84 | 2.43 | 2.56 | **3.23** | 2.90 | 2.90 | **4.33** |

Table 4: Illustrative Argument pairs, along with the predicted scores from individual feature sets, predicted(**AFS**) and the Mechanical Turk human topline (**MT AFS**). The best performing feature set is shown in bold. GC=Gun Control, DP=Death Penalty, GM=Gay Marriage.

nation for learning similarity with vector representations works much better than the common practice of reducing a pair of vectors to a single score using cosine similarity. Previous work (Li et al., 2015; Pennington et al., 2014) also shows that all dimensions are not equally useful predictors for a specific task. For sentiment classification, Li et al. (2015) find that "too large a dimensionality leads many dimensions to be non-functional ... causing two sentences of opposite sentiment to differ only in a few dimensions." This may also be the situation for the 300-dimensional embeddings used for AFS. Hence, when using concatenation, single dimensions can be weighted to adjust for non-functional dimensions, but using cosine makes this per-dimension weighting impossible. This might explain why our custom word2vec model outperforms the Google model when using cosine as compared to concatenation, i.e. more dimensions are informative in the custom model, but overall, the Google model provides more complementary information when non-functional dimensions are accounted for. More analysis is needed to fully support this claim.

To qualitatively illustrate some of the differences between our final AFS regressor model (Row 8 of Table 3) and several baselines, we apply the model to a set-aside 200 pairs per topic. Table 4 shows examples selected to highlight the strengths of AFS prediction for different models as compared to the AFS gold standard scores.

MT AFS values near 1 indicate same topic but no similarity. Rows GC1 and DP2 talk about totally different facets and only share the same topic (AFS = 1). Rouge and Ngram features based on word overlap predict scores that are too high. In contrast, LIWC dependencies and word2vec based on concept and semantic overlap are more accurate. MT values near 3 indicate same facet but somewhat different arguments. Arguments in row GM4 talk about marriage rights to all, and there is some overlap in these arguments beyond simply being the same topic, however the speakers are on opposite stance sides. Both of the arguments in row GM5 (MT AFS of 3.3) reference the same facet of the financial and legal benefits available to married couples, but Arg2 is more specific. Both Word2vec and our trained AFS model can recognize the similarity in the concepts in the two arguments and make good predictions.

MT values above 4 indicate two arguments that are the same facet and very similar. Row DP6 gets a high Rouge overlap score and Word2vec relates 'lower crime rate' as semantically similar to 'deter murder rates' thus yielding an accurately high AFS score. DP7 is an example where LIWC dependencies perform better as compared to other features, because it focuses in on the dependency between the death penalty and cost, but none of the models do well at predicting the MT AFS score. One issue here may be that, despite our attempts to sample pairs with more representatives of high AFS, there is just less training data available for this part of the distribution. Hence all the regressors will be conservative at predicting the highest values. We hope in future work to improve our AFS regressor by finding additional methods for populating the training data with more highly similar pairs.

## 4 Related Work

There are many theories of argumentation that might be applicable for our task (Jackson and Jacobs, 1980; Reed and Rowe, 2004; Walton et al., 2008; Gilbert, 1997; Toulmin, 1958; Dung, 1995), but one definition of argument structure may not work for every NLP task. Social media arguments are often informal, and do not necessarily follow logical rules or schemas of argumentation (Stab and Gurevych, 2014; Peldszus and Stede, 2013; Ghosh et al., 2014; Habernal et al., 2014; Goudas et al., 2014; Cabrio and Villata, 2012).

Moreover, in social media, segments of text that are argumentative must first be identified, as in our **Task1**. Habernal and Gurevych (2016) train a classifier to recognize text segments that are argumentative, but much previous work does Task1 manually. Goudas et al. (2014) annotate 16,000 sentences from social media documents and consider 760 of them to be argumentative. Hasan and Ng (2014) also manually identify argumentative sentences, while Boltuzic and Šnajder (2014) treat the whole post as argumentative, after manually removing "spam" posts. Biran and Rambow (2011) automatically identify justifications as a structural component of an argument.

Other work groups semantically-similar classes of **reasons** or **frames** that underlie a particular speaker's stance, what we call ARGUMENT FACETS. One approach categorizes sentences or posts using topic-specific argument labels, which are functionally similar to our facets as discussed above (Conrad et al., 2012; Hasan and Ng, 2014; Boltuzic and Šnajder, 2014; Naderi and Hirst, 2015). For example, Fig. 2 lists facets **A1** to **A8** for Gun Control from the IDebate website; Boltuzic and Šnajder (2015) use this list to label posts. They apply unsupervised clustering using a semantic textual similarity tool, but evaluate clusters using their hand-labelled argument tags. Our method instead explicitly models graded similarity of sentential arguments.

## 5 Conclusion and Future Work

We present a method for scoring argument facet similarity in online debates using a combination of hand-engineered and unsupervised features with a correlation averaging 0.63 compared to a human top line averaging 0.68. Our approach differs from similar work that finds and groups the "reasons" underlying a speakers stance, because our models are based on the belief that it is not possible to define a finite set of discrete facets for a topic. A qualitative analysis of our results, illustrated by Table 4, suggests that treating facet discovery as a similarity problem is productive, i.e. examination of particular pairs suggests facets about legal and financial benefits for same-sex couples, the claim that the death penalty does not actually affect murder rates, and an assertion that "they", implying "congress", do not have the express, enumerated power to pass legislation restricting guns.

Previous work shows that metrics used for evaluating machine translation quality perform well on paraphrase recognition tasks (Madnani et al., 2012). In our experiments, ROUGE performed very well, suggesting that other machine translation metrics such as Terp and Meteor may be useful (Snover et al., 2009; Lavie and Denkowski, 2009). We will explore this in future work.

In future, we will use our AFS regressor to cluster and group similar arguments and produce *argument facet summaries* as a final output of our pipeline. Habernal and Gurevych (2015) apply clustering in argument mining by averaging word embeddings from posts and sentences from debate portals, clustering the resulting averaged vectors, and then computing distance measures from clusters to unseen sentences ("classification units") as features. Cosine similarity between weighted and summed vector representations is also a common approach, and Boltuzic and Šnajder (2015) show word2vec cosine similarity beats bag-of-words and STS baselines when used with clustering for argument identification.

Finally, our AQ extractor treats all posts on a topic equally, operating on a set of concatenated posts. We will explore other sampling methods to ensure that the AQ extractor does not eliminate arguments made by less articulate citizens, by e.g. enforcing that *"Every speaker in a debate contributes at least one argument"*. We will also sample by stance-side, so that summaries can be organized using "Pro" and "Con", as in curated summaries. Our final goal is to combine quality-based argument extraction, our AFS model, stance, post and author level information, so that our summaries represent the diversity of views on a topic, a quality not always guaranteed by summarization techniques, human or machine.

## Acknowledgments

This work was supported by NSF CISE RI 1302668. Thanks to Keshav Mathur and the three anonymous reviewers for helpful comments.

## A  Appendix

| Score | Scoring Criteria |
|---|---|
| 3 | The phrase is clearly interpretable AND either expresses an argument, or a premise or a conclusion that can be used in an argument about a facet or a sub-issue for the topic of gay marriage. |
| 2 | The phrase is clearly interpretable BUT does not seem to be a part of an argument about a facet or a sub-issue for the topic of gay marriage. |
| 1 | The phrase cannot be interpreted as an argument. |

Figure 8: Argument Quality HIT as instantiated for the topic Gay Marriage.

Figure 8 shows the definitions used in our Argument Quality HIT.

Figure 9 shows the relation between predicted AQ score and gold-standard argument quality annotations.

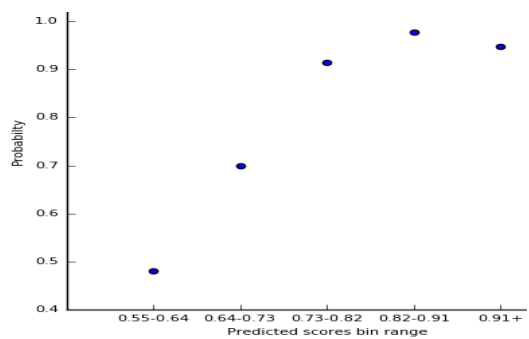

Figure 9: Probability of sentential argument for AQ score across bin for Death Penalty.

Figure 10 provides our definition of FACET and instructions for AFS annotation. This is repeated here from (Misra et al., 2015) for the reader's convenience.

> **Facet**: A facet is a low level issue that often reoccurs in many arguments in support of the author's stance or in attacking the other author's position. There are many ways to argue for your stance on a topic. For example, in a discussion about the death penalty you may argue in favor of it by claiming that it deters crime. Alternatively, you may argue in favor of the death penalty because it gives victims of the crimes closure. On the other hand you may argue against the death penalty because some innocent people will be wrongfully executed or because it is a cruel and unusual punishment. Each of these specific points is a facet.
> For two utterances to be about the same facet, it is not necessary that the authors have the same belief toward the facet. For example, one author may believe that the death penalty is a cruel and unusual punishment while the other one attacks that position. However, in order to attack that position they must be discussing the same facet.

> We would like you to classify each of the following sets of pairs based on your perception of how SIMILAR the arguments are, on the following scale, examples follow.
> (5) Completely equivalent, mean pretty much exactly the same thing, using different words.
> (4) Mostly equivalent, but some unimportant details differ. One argument may be more specific than another or include a relatively unimportant extra fact.
> (3) Roughly equivalent, but some important information differs or is missing. This includes cases where the argument is about the same FACET but the authors have different stances on that facet.
> (2) Not equivalent, but share some details. For example, talking about the same entities but making different arguments (different facets)
> (1) Not equivalent, but are on same topic
> (0) On a different topic

Figure 10: Definitions used for Facet and AFS in MT HIT.